\documentclass[11pt,a4]{article}

\usepackage{graphics,amsfonts,latexsym,a4wide}

\usepackage{graphicx,url, color, bm}
\newcommand{\est}{{\hfill $\star$}}

\newcommand{\tran}{^{\top}}

\newcommand{\diag}{\mbox {\rm diag}}

\newcommand{\prob}{\mbox{\rm Prob}}
\newcommand{\med}{\mbox{\rm med}}

\newcommand{\yvet}{{\mathbf y}}

\newcommand{\inv}{^{-1}}
\newcommand{\beq}{\begin{equation}}
\newcommand{\eeq}{\end{equation}}
\newcommand{\bea}{\begin{eqnarray}}
\newcommand{\eea}{\end{eqnarray}}
\newcommand{\beas}{\begin{eqnarray*}}
\newcommand{\eeas}{\end{eqnarray*}}
\newcommand{\ba}{\begin{array}}
\newcommand{\ea}{\end{array}}
\newcommand{\bit}{\begin{itemize}}
\newcommand{\eit}{\end{itemize}}
\newcommand{\ben}{\begin{enumerate}}
\newcommand{\een}{\end{enumerate}}
\newcommand{\dss}{\displaystyle}

\newcommand{\Real}[1]{ { {\mathbb R}^{#1} } }

\newtheorem{definition}{Definition}
\newtheorem{proposition}{Proposition}
\newtheorem{remark}{Remark}

\newcommand{\calD}{{\cal D}}
\newcommand{\calJ}{{\cal J}}

\newcommand{\calE}{{\cal E}}

\newcommand{\E}{\mathbb{E}}

\newcommand{\qed}{\hfill $\square$}

\usepackage{amsmath}
\usepackage[mathscr]{eucal}

\begin{document}
\author{Giuseppe C. Calafiore and Giulia Fracastoro
\thanks{Dipartimento di Elettronica e Telecomunicazioni,
Politecnico di Torino, Italy.
Tel.: +39-011-564.7071;  E-mail: {\tt giuseppe.calafiore@polito.it} }
}
\title{Sparse $\ell_1$ and $\ell_2$ Center Classifiers\footnote{This research was funded in part by sumup.ai.}}
\date{}
\maketitle

\begin{abstract}
The nearest-centroid classifier is a simple linear-time classifier based on computing the centroids of the data classes in the training phase, and then assigning a new datum to the class corresponding to its nearest centroid. Thanks to its very low computational cost, the  nearest-centroid classifier is still widely used in machine learning, despite the development of many other more sophisticated classification methods. In this paper, we propose two sparse variants of the nearest-centroid  classifier, based respectively  on  $\ell_1$ and $\ell_2$ distance criteria. The proposed sparse classifiers perform simultaneous classification and feature selection, by detecting the features that are most relevant for the classification purpose. We show that training of the proposed sparse models, with both distance criteria, can be performed exactly (i.e., the globally optimal set of features is selected) and at a quasi-linear computational cost.
The experimental results show that the proposed methods are competitive in accuracy with state-of-the-art feature selection techniques, while having a significantly lower computational cost.
\end{abstract}

\section{Introduction}
In the last years, the technological development has led to a massive proliferation of large-scale datasets. The processing of these large amounts of data poses many new challenges and there is a strong need of algorithms that mildly scales (e.g., linearly or quasi-linearly) with the dataset size. For this reason, classification methods with a very low computational cost, such as Naive Bayes and the nearest centroid classifier, are an appealing choice in this endeavour. Sometimes these methods are the only feasible approaches, since more sophisticated techniques would be too slow.

When the number of features in a datasets is very high, feature selection is a necessary step  of any machine learning algorithm. Feature selection consists in selecting a subset of features of the dataset, choosing the most relevant ones. Besides reducing the dataset size, feature selection has some other important advantages. First, it eliminates noisy or irrelevant features, reducing the risk of overfitting. Second, by selecting only the most significant features, it improves the interpretability of the model. State-of-the-art feature selection methods are usually based on some heuristics without any guarantee of optimality. 
Some of them, such as LASSO \cite{tibshirani1996regression} or $\ell_1$-regularized logistic regression \cite{ng2004feature}, are based on a convex optimization problem with a $\ell_1$-norm penalty on the regression coefficients to promote sparsity. The main drawback of these techniques is that they are usually computationally expensive. Other methods, such as Odds Ratio \cite{mladenic1999feature}, propose a different approch, performing a feature ranking based on their inherent characteristics. These methods are usually very fast, but often their performance in terms of accuracy is very poor. Recently, \cite{askari2019naive} have presented a feature selection method targeted for a Naive Bayes classifier. This method can provide an optimal solution in the case of binary data, and an approximate upper bound for general data. 

In this paper, we propose a sparse centroid classifier. The proposed method can simultaneously perform feature selection and classification. We introduce two different variants of the method, namely $\ell_1$-sparse centroids and $\ell_2$-sparse centroids, where we consider the $\ell_1$ and the $\ell_2$ distance criteria, respectively. The $\ell_2$ case is a sparse variant of the nearest centroids classifier \cite{han2000centroid}, which is a widely used classifier especially in text classification. Instead, the $\ell_1$ case is related to the median classifier \cite{hall2009median}, that is more robust to noise than the nearest centroid classifier. We prove that both the proposed method can select the optimal subset of features for the corresponding  classifier. The experimental results show that the proposed techniques achieves similar performance as state-of-the-art feature selection methods, but with a significantly lower computational cost.

\section{Preliminaries on center-based classifiers}
\label{sec:prelim}
Let
\beq
X = \left[x^{(1)} \,\cdots\, x^{(n)} \right] \in\Real{m,n},
	\label{eq:Xdata}
	\eeq
be a given data matrix whose columns $x^{(j)}\in\Real{m}$, $j=1,\ldots,n$, contain feature vectors from $n$ observations, and  let 
$\yvet\in\Real{n}$ be a given vector such that $y_j\in\{-1, +1\}$ is the class label  corresponding to the $j$-th observation.
We consider a binary classification problem, in which  a new  observation vector $x\in\Real{m}$
is to be assigned to the positive class $C_+$ (corresponding to $y = +1$) or to the negative class $C_-$ (corresponding to $y=-1$). 
To this purpose, the {\em nearest centroid classifier}  \cite{han2000centroid, manning2008vector,tibshirani2002diagnosis} is a well-known classification model, which works by assigning the class label based on the least Euclidean distance from $x$ to the centroids of the classes. The centroids
are computed on the basis of the training data as
\beq
\bar x^+ = \frac{1}{n_+} \sum_{j\in\calJ^+} x^{(j)}, \quad
\bar x^- = \frac{1}{n_-} \sum_{j\in\calJ^-} x^{(j)},
\label{eq:centroids}
\eeq
where $\calJ^+ \doteq \{j\in\{1,\ldots,n\}:\, y_j = +1\}$ contains the indices of the observations in the positive class,
$\calJ^- \doteq \{j\in\{1,\ldots,n\}:\, y_j = -1\}$ contains the indices of the observations in the negative class, and
$n_+$, $n_-$ are the cardinalities of $\calJ^+$ and $\calJ^-$, respectively.
A new observation vector $x$ is classified as positive or negative according to the sign of
\[
\Delta_2(x) = \|x- \bar x^- \|_2^2 - \|x- \bar x^+\|_2^2 ,
	\label{eq:discr}
\]
that is, $x$ is classified in the positive class if its Euclidean distance from the positive centroid is smaller that its distance from the negative centroid, and viceversa for the negative class.
The discrimination surface for the centroid classifier is linear with respect to  $x$, since
\bea
\Delta_2(x) &=& \|x\|_2^2 +   \|\bar x^- \|_2^2  - 2 x\tran \bar x^-
-  \|x\|_2^2 -   \|\bar x^+ \|_2^2  +2 x\tran \bar x^+ \nonumber  \\
&=& (\|\bar x^- \|_2^2 -  \|\bar x^+ \|_2^2) + 2 x\tran (\bar x^+ - \bar x^-),
\label{eq:discrlin}
\eea
where the coefficient in the linear  term of the classifier is given by vector $w \doteq \bar x^+ - \bar x^-$.
Notice   that, whenever $\bar x^+_i = \bar x^-_i$ for some component $i$ (i.e., $w_i=0$), the corresponding feature $x_i$ in $x$ is irrelevant for the purpose of classification. 

\begin{remark} \rm
We observe that the centroids in \eqref{eq:centroids} can be seen as the optimal solutions to the following optimization problem:
\beq
\min_{\theta^+,\theta^-\in\Real{m}}\, \frac{1}{n_+}\sum_{j\in\calJ^+}  \| x^{(j)} - \theta^+\|_2^2
+\frac{1}{n_-}\sum_{j\in\calJ^-}  \| x^{(j)} - \theta^-\|_2^2.
\label{eq:centroidseq}\eeq
That is, the centroids are the points that minimize the average squared distance to the samples within each class.
A proof of this fact is immediate, by taking the gradient of the objective in \eqref{eq:centroidseq} with respect to\ $\theta^+$ and equating it to zero, and then doing the same thing for $\theta^-$. The two problems are actually decoupled, so the two coefficients 
$ \frac{1}{n_+}$ and $ \frac{1}{n_-}$ play no role here in terms of the optimal solution. However, they have been introduced for balancing the contribution of the residuals of the two classes. \est
\end{remark}

We shall call \eqref{eq:centroidseq} the (plain) $\ell_2$-center classifier training problem, and $\Delta_2$ in \eqref{eq:discrlin} the corresponding discrimination function. The usual centroids in \eqref{eq:centroids} are thus the points that minimize the average $\ell_2$ distance from the respective class representatives. This interpretation opens the way to considering different types of metrics for computing centers. In particular, there exist an extensive literature on the favorable properties of the $\ell_1$ norm criterion, which is well known to provide center estimates that are robust to outliers. The natural $\ell_1$ version of problem \eqref{eq:centroidseq} is
\beq
\min_{\theta^+,\theta^-\in\Real{m}}\, \frac{1}{n_+}\sum_{j\in\calJ^+}  \| x^{(j)} - \theta^+\|_1
+\frac{1}{n_-}\sum_{j\in\calJ^-}  \| x^{(j)} - \theta^-\|_1,
\label{eq:centroidseqL1}\eeq
which we shall call  the (plain) $\ell_1$-center classifier training problem. It is known that an optimal solution to problem \eqref{eq:centroidseqL1} is obtained, for each $i=1,\ldots,m$, by taking $\theta^+_i$ to be the {\em median} of the values
$x^{(j)}_i$ in the positive class, and $\theta^-_i$ to be the { median} of the values
$x^{(j)}_i$ in the negative class, see also the more general result given in Proposition~\ref{prop:wmedian}. We let
\beq
\mu^+ \doteq \med(\{x^{(j)}\}_{j\in\calJ^+}), \quad
\mu^- \doteq \med(\{x^{(j)}\}_{j\in\calJ^-}),
\label{eq:medianv}
\eeq
where $\med$ computes the median of its input vector sequence along each component, i.e., for each $i=1,\ldots,m$,  $\mu^+_i$ is the median of
$\{x_i^{(j)}\}_{j\in\calJ^+}$, and $\mu^-_i$ is the median of
$\{x_i^{(j)}\}_{j\in\calJ^-}$.
The classification in the $\ell_1$-center classifier is made by computing the distances from the new datum $x$ and the
$\ell_1$ centers of the classes, and assigning $x$ to the closest center, that is, we compute
\[
\Delta_1(x) \doteq 
\|x-  \mu^- \|_1- \|x- \mu^+\|_1 ,
	\label{eq:discrL1}
\]
and assign $x$ to the positive or negative class depending on the sign of $\Delta_1(x) $.
We observe that, contrary to the $\ell_2$ case, the discrimination criterion based on the sign of $\Delta_1(x) $
 is not linear in $x$. However, expressed more explicitly in its components, $\Delta_1(x) $ is written as
\[
\Delta_1(x) = \sum_{i=1}^m
\left( |x_i-  \mu^-_i |- |x_i- \mu^+_i| \right),
	\label{eq:discrL1i}
\]
and we observe again, like in the  $\ell_2$ case, that the contribution to $ \Delta_1(x)$ from the $i$th feature $x_i$ is identically zero when $\mu^-_i = \mu^+_i$.

\section{Sparse $\ell_1$ and $\ell_2$ center classifiers}
\label{sec:sparseclass}
In Section~\ref{sec:prelim} we observed that, for both the $\ell_2$ and the $\ell_1$ distance criteria, 
the discrimination is insensitive to the $i$th feature whenever $\theta^+_i-\theta^-_i = 0$, where 
$\theta^+$, $\theta^-$ are the two class centers. The {\em sparse} classifiers that we introduce in this section are aimed precisely at computing optimal class centers such that the center difference 
$\theta^+ - \theta^-$
is $k$-sparse, meaning that $\|\theta^+ - \theta^-\|_0 \leq k$, where $\|\cdot\|_0$ denotes the number of nonzero entries (i.e., the cardinality) of
its argument, and $k\leq m$ is a given cardinality bound. Such type of sparse classifiers will thus perform simultaneous classification and { feature selection},  by detecting  which $k$ out of the total $m$ features are relevant for the classification purposes. 
We next formally define the sparse $\ell_2$ and $\ell_1$ center classifier training problems.

\begin{definition}[Sparse $\ell_2$-center classifier]
A sparse $\ell_2$-center classifier is a model which classifies an input feature vector $x\in\Real{m}$ into a positive or a negative class, according to the sign of the discrimination function
\bea
\Delta_2(x) &=& \|x - \theta^- \|_2^2 -  \|x - \theta^+ \|_2^2		\nonumber \\
&=& (\|\theta^- \|_2^2 -  \|\theta^+ \|_2^2) + 2 x\tran (\theta^+ - \theta^-),
\label{eq:discrlinL2}
\eea
where the sparse $\ell_2$-centers $\theta^+$,  $\theta^-$ are learned from a data batch \eqref{eq:Xdata} as the optimal solutions of the problem
\bea
\min_{\theta^+,\theta^-\in\Real{m}} & \frac{1}{n_+}\sum_{j\in\calJ^+}  \| x^{(j)} - \theta^+\|_2^2 
+\frac{1}{n_-}\sum_{j\in\calJ^-}  \| x^{(j)} - \theta^-\|_2^2 \label{eq:L2obj} \\
\mbox{subject to:} & \| \theta^+ - \theta^-\|_0 \leq k,  \nonumber \label{eq:L2sparsity}
\eea
where $k\leq m$ is a given upper bound on the cardinality of $\theta^+ - \theta^-$.
\end{definition}

\begin{definition}[Sparse $\ell_1$-center classifier]
A sparse $\ell_1$-center classifier is a model which classifies an input feature vector $x\in\Real{m}$ into a positive or a negative class, according to the sign of the discrimination function
\bea
\Delta_1(x) \doteq 
\|x-  \theta^- \|_1- \|x- \theta^+ \|_1 ,
\eea
where the sparse $\ell_1$-centers $\theta^+$,  $\theta^-$ are learned from a data batch \eqref{eq:Xdata} as the optimal solutions of the problem
\bea
\min_{\theta^+,\theta^-\in\Real{m}} & \frac{1}{n_+}\sum_{j\in\calJ^+}  \| x^{(j)} - \theta^+\|_1 
+\frac{1}{n_-}\sum_{j\in\calJ^-}  \| x^{(j)} - \theta^-\|_1 \label{eq:L1obj} \\
\mbox{subject to:} & \| \theta^+ - \theta^-\|_0 \leq k, \nonumber  \label{eq:L1sparsity}
\eea
where $k\leq m$ is a given upper bound on the cardinality of $\theta^+ - \theta^-$.
\end{definition}

A perhaps notable fact is that both 
the sparse $\ell_2$  and the sparse $\ell_1$ classifier training problems can be solved {exactly} and with  almost-linear-time complexity (this fact is proved  in the next sections), which also makes them good candidates for efficient feature selection methods in two-phase (feature selection + actual classifier training) classifier training  procedures.

\section{Training the sparse $\ell_2$-center classifier}
\label{sec:L2train}
We next discuss how to solve the training problem in \eqref{eq:L2obj}.
Let us denote by $J$ the objective to be minimized in   \eqref{eq:L2obj}. By expanding the squares and using \eqref{eq:centroids}, we have
\beas
J &=& 
\frac{1}{n_+}\sum_{j\in\calJ^+}  \| x^{(j)} \|_2^2 
+ \frac{1}{n_-}\sum_{j\in\calJ^-}  \| x^{(j)} \|_2^2 +
 \| \theta ^+ \|_2^2 +   \| \theta ^- \|_2^2 - 2 \bar x^{+\top} \theta^+ - 2  \bar x^{-\top} \theta^- \nonumber \\
 &=& \mbox{cost.} +
 \| \theta ^+ \|_2^2 +   \| \theta ^- \|_2^2 - 2 \bar x^{+\top} \theta^+ - 2  \bar x^{-\top} \theta^-.
\eeas
Let now $\calE$ denote a fixed set of indices of cardinality $m-k$, and $\calD$ denote the complementary set, that is, 
$\calD = \{1,\ldots,m\} \setminus  \calE$. For any vector $x\in\Real{m}$ we next use the notation $x_\calD$ to denote a vector of the same dimension as $x$ which coincides with $x$ at the locations in $\calD$ and it is zero elsewhere. We define analogously 
$x_{\calE}$, so that $x = x_\calD + x_{ \calE}$. We then let
\beas
\theta ^+ &=& \theta ^+_\calD +  \theta^+_{\calE} \\
\theta ^- &=& \theta ^-_\calD +  \theta^-_{\calE} .
\eeas
Suppose that we fixed the set $\calE$ of the indices where $ \theta^+ - \theta^- $ is zero  (we shall discuss later how to eventually optimize over this choice of the index set), so that $ \theta^+_\calE - \theta^-_\calE = 0$. We can therefore set
\[
 \theta^+_{\calE} = \theta^-_\calE \doteq \theta_\calE,
\]
whence
\beas
\theta ^+ &=& \theta ^+_\calD +  \theta_\calE \\
\theta ^- &=& \theta ^-_\calD +  \theta_\calE .
\eeas
With such given choice of the zero index set, and using the above expressions for $\theta ^+,\theta ^-$, the problem objective becomes
\beas
J_\calE &=& 
 \mbox{cost.} +
 \| \theta ^+ \|_2^2 +   \| \theta ^- \|_2^2 - 2 \bar x^{+\top} \theta^+ - 2  \bar x^{-\top} \theta^- \nonumber \\
 &=& 
 \mbox{cost.} +
2 \| \theta_\calE\|_2^2  -4\tilde x\tran \theta_\calE   + \|  \theta ^+_\calD\|_2^2 +  
 \|  \theta ^-_\calD \|_2^2  - 2\bar x^{+\top} \theta^+_\calD   - 2\bar x^{-\top} \theta^-_\calD   ,
 \eeas
 where we defined
 \beq
 \tilde x \doteq \frac{\bar x^+ + \bar x^-}{2}.
 \label{eq:midpoint}
 \eeq
For given zero index set $\calE$ we can therefore minimize $J_\calE$ with respect to 
$\theta_\calE$, $\theta^+_\calD$, and $\theta^-_\calD$. By simply equating the respective gradients to zero, we obtain that the optimal parameter values are
\[
\theta_\calE^* = \tilde x_\calE, \quad 
\theta^{+*}_\calD = \bar x^+_\calD, \quad 
\theta^{-*}_\calD = \bar x^-_\calD. 
\]
Substituting these optimal values back into $J_\calE$ we obtain
\beas
J_\calE^* &=& 
 \mbox{cost.} 
 - 2 \|\tilde x_\calE \|_2^2 - \| \bar x^+_\calD \|_2^2 - \| \bar x^-_\calD \|_2^2 \nonumber \\
 &=& 
 \mbox{cost.} 
 -\frac{1}{ 2} \|\bar x^+_\calE + \bar x^-_\calE \|_2^2 - \| \bar x^+_\calD \|_2^2 - \| \bar x^-_\calD \|_2^2 \nonumber \\
 &=& 
 \mbox{cost.} 
 -\frac{1}{ 2} \|\bar x^+_\calE \|_2^2 
 -\frac{1}{ 2} \|\bar x^-_\calE \|_2^2 
 - \bar x^{+\top}_\calE \bar x^-_\calE
 - \| \bar x^+_\calD \|_2^2 - \| \bar x^-_\calD \|_2^2 \nonumber \\
  &=& 
 \mbox{cost.} 
 -\frac{1}{ 2}( \|\bar x^+_\calE \|_2^2  +\| \bar x^+_\calD \|_2^2)
 -\frac{1}{ 2}( \|\bar x^-_\calE \|_2^2 +\| \bar x^-_\calD \|_2^2)
 - \bar x^{+\top}_\calE \bar x^-_\calE
 - \frac{1}{2}(\| \bar x^+_\calD \|_2^2 +   \| \bar x^-_\calD \|_2^2 ) \nonumber  \\
  &=& 
 \mbox{cost.} 
 -\frac{1}{ 2}\|\bar x^+ \|_2^2 
 -\frac{1}{ 2} \|\bar x^- \|_2^2
 - \bar x^{+\top}_\calE \bar x^-_\calE
 - \frac{1}{2}( \| \bar x^+_\calD  - \bar x^-_\calD \|_2^2 
 + 2  \bar x^{+\top}_\calD \bar x^-_\calD) \nonumber  \\
  &=& 
 \mbox{cost.} 
 -\frac{1}{ 2}\|\bar x^+ \|_2^2 
 -\frac{1}{ 2} \|\bar x^- \|_2^2
 - (\bar x^{+\top}_\calE \bar x^-_\calE + \bar x^{+\top}_\calD \bar x^-_\calD)
 - \frac{1}{2} \| \bar x^+_\calD  - \bar x^-_\calD \|_2^2  \nonumber  \\
  &=& 
 \mbox{cost.} 
 -\frac{1}{ 2}\|\bar x^+ \|_2^2 
 -\frac{1}{ 2} \|\bar x^- \|_2^2 - 
 \bar x^{+\top} \bar x^-
 - \frac{1}{2} \| \bar x^+_\calD  - \bar x^-_\calD \|_2^2  \nonumber  \\
  &=& 
 \mbox{cost.} 
 -\frac{1}{ 2}\|\bar x^+   + \bar x^-  \|_2^2 
 - \frac{1}{2} \| \bar x^+_\calD  - \bar x^-_\calD \|_2^2  . \label{eq:Jest}
 \eeas
 This last expression shows that $J_\calE^*$ depends on the choice of the zero index set $\calE$ only via
 the term $ \| \bar x^+_\calD  - \bar x^-_\calD \|_2^2$ involving the complementary set $\calD$.
 Minimizing $J_\calE^*$ with respect to the index set $\calE$ thus amounts to maximizing $ \| \bar x^+_\calD  - \bar x^-_\calD \|_2^2$
 with respect to the complementary index set $\calD$, that is
 \[
J^* =  \mbox{cost.}'  -  \frac{1}{2}  \max_{|\calD| \leq  k} \,  \| \bar x^+_\calD  - \bar x^-_\calD \|_2^2.
\label{eq:Jstar}
 \]
 The solution to this problem is immediate: we construct the difference vector $\delta \doteq  \bar x^+  - \bar x^-$ and let 
 $\calD^*$ contain the indices of the $k$ largest elements of $|\delta|$.
 We have therefore proved the following
 
 \begin{proposition}
  \label{prop:ell2opt}
 The optimal solution of problem \eqref{eq:L2obj} is obtained as follows:
 \ben
 \item Compute the standard class centroids $\bar x^+$, $\bar x^-$ according to \eqref{eq:centroids};
 \item Compute the centroids midpoint $\tilde x$ according to \eqref{eq:midpoint}, and the centroids difference  $\delta \doteq  \bar x^+  - \bar x^-$;
 \item Let $\calD$ be the set of the indices of the $k$ largest absolute value elements in vector $\delta $, and let $\calE$ be the complementary index set;
 \item The optimal parameters $\theta ^+$, $\theta^-$ are given by
\beas
\theta ^+ &=& \bar x^+_\calD +  \tilde x_\calE \\
\theta ^- &=&\bar x^-_\calD +  \tilde x_\calE .
\eeas
 \een
 \end{proposition}

\begin{remark}[Numerical complexity for training the sparse $\ell_2$ classifier]\rm 
Steps 1-2 in \\ Proposition~\ref{prop:ell2opt} essentially require computing $mn$ sums. Finding the $k$ largest elements in 
Step 3 takes $O(m\log k)$ operations (using, e.g.,  min-heap sorting), whence the whole procedure takes $O(mn) + O(m\log k)$ operations. 
Thus, while training a plain centroid classifier takes $O(mn)$ operations (which, incidentally, is also the complexity figure for training  a classical Naive Bayes classifier), adding exact
sparsity  comes at the quite moderate extra cost of $O(m\log k)$ operations. \est
\end{remark}

\begin{remark}[Online recursive training]\rm 
The sparse  $\ell_2$ center classifier training procedure
is amenable to efficient online implementation, since the class centers are easily updatable as soon as new data comes in. 
Denote 
by $\bar x(\nu)$ 
the centroid of one of the two classes when $\nu$ observations $\xi^{(1)},\ldots,\xi^{(\nu)}$ in that class are present:
$
\bar x(\nu) = \frac{1}{\nu} \sum_{j=1}^\nu \xi^{(j)}
$.
If a new observation $\xi^{(\nu+1)}$ in the same class becomes available, the new centroid will be
\beas
\bar x(\nu+1) &=& \frac{1}{\nu+1} \sum_{j=1}^{\nu+1} \xi^{(j)} =
 \frac{1}{\nu+1} \left( \sum_{j=1}^{\nu} \xi^{(j)} + \xi^{(\nu+1)} \right)\\
 &=& 
 \frac{\nu}{\nu+1} \bar x(\nu) +  \frac{1}{\nu+1} \xi^{(\nu+1)}.
\eeas
This latter formula gives the new centroid as a weighted linear combination of the previous centroid and of the new observation. An online version of the procedure in Proposition~\ref{prop:ell2opt} is thus readily obtained, in which only the current 
centroids are kept into memory and, 
as soon as a new datum is available, the corresponding centroid is updated (this takes $O(m)$ operations, or less if the datum is sparse) and the feature ranking is recomputed 
(this takes $O(m\log k)$ operations).  A sparse  $\ell_2$ center classifier can therefore be trained online with $O(m)$ memory storage and $O(m\log k)$ operations per update.
 \est
\end{remark}

\begin{remark}[Sparsity-accuracy tradeoff]\rm 
\label{rem:spacc}
As it is customary with sparse methods,
 in practice a whole sequence of training problems is solved at different levels of sparsity, say from $k=1$ (only one feature selected) to $k=m$ (all features selected), accuracy is evaluated for each model via cross validation, and then the resulting sparsity-accuracy tradeoff
 curve is examined for the purpose of selection of the most suitable $k$ level. Most feature selection methods, including sparse SVM, the Lasso \cite{tibshirani1996regression}, and the sparse Naive Bayes method \cite{askari2019naive}, require repeatedly solving the training problem for each $k$, albeit typically warm-starting the optimization procedure with the solution from the  previous $k$ value.
In the sparse $\ell_2$ classifier, instead, one can fully order the vector $|\bar x^+ -\bar x^-|$ 
 only once, at a computational cost of $O(m\log m)$, and then the optimal solutions  are obtained, for any $k$, by simply selecting
 in Step 3 of Proposition~\ref{prop:ell2opt} the first $k$ elements of the ordered vector.
 \est
\end{remark}

\subsection{Mahalanobis  distance classifier}
\label{sec:mahala}
A variant of the $\ell_2$ centroid classifier is obtained by considering the Mahalanobis distance instead of the Euclidean distance.
Letting $S$ denote an estimated data covariance matrix, the Mahalanobis
distance from a point $z$ to a center $\theta^{\pm}$ is defined by
\[
\mbox{dist}_S(z, \theta^{\pm}) = (z-\theta^{\pm})\tran S\inv (z-\theta^{\pm}).
\] 
This leads to the  Mahalanobis training problem
\beas
\min_{\theta^+,\theta^-\in\Real{m}} & 
\frac{1}{n_+}\sum_{j\in\calJ_+}  (x^{(j)} - \theta^+)\tran S\inv (x^{(j)} - \theta^+)
+\frac{1}{n_-}\sum_{j\in\calJ_-}   (x^{(j)} - \theta^-)\tran S\inv (x^{(j)} - \theta^-) 
\eeas
Classification of a new observation $x$ in this setting is  performed according to the sign of
\bea
\Delta_M(x) &=& 
 (x - \theta^-)\tran S\inv (x - \theta^-)
 - 
 (x - \theta^+)\tran S\inv (x - \theta^+)
  \nonumber\\
&=&
 ( \theta^-  S\inv \theta^-  - \theta^+  S\inv \theta^+ ) 
 + 2(  \theta^+ -  \theta^-)\tran S\inv x.
\label{eq:discrlinspM}
\eea
By introducing 
 a change of variables of the type 
\[
\xi^{(j)} \doteq  S^{-1/2} x^{(j)},\; j=1,\ldots,n;\quad
\omega^{\pm} \doteq S^{-1/2} \theta^\pm,
\]
where $S^{-1/2}$ is the matrix square root of  $S\inv$,
we see that the Mahalanobis training problem, in the new variables,  becomes
\bea
\min_{\omega^+,\omega^-\in\Real{m}} & 
\frac{1}{n_+}\sum_{j\in\calJ_+}  \| \xi^{(j)} - \omega^+ \|_2^2 
+\frac{1}{n_-}\sum_{j\in\calJ_-}   \| \xi^{(j)} - \omega^- \|_2^2
\label{eq:centroidseqM} \eea
and the discrimination function, for $\xi = S^{-1/2}  x$, becomes
\beas
\Delta_M(\xi) &=& 
(\| \omega^-\|_2^2 - \| \omega^+\|_2^2)
 +  2(  \omega^+ - \omega^-)\tran \xi.
\label{eq:discrlinspMomega}
\eeas
Problem \eqref{eq:centroidseqM} is now a standard $\ell_2$ center classifier problem, hence its sparse version can be readily solved by means of the algorithm outlined in Proposition~\ref{prop:ell2opt}.
It should however be observed that in this case one obtains sparsity in the transformed center difference $ \omega^+ - \omega^-$, which implies a selection of the transformed features  in $\xi = S^{-1/2}  x$. 
One relevant special case arises when $S = \diag(\sigma_1^2,\ldots,\sigma_m^2)$, in which case
the data transformation $\xi = S^{-1/2}  x$ simply  amounts to normalizing each feature $x_i$  by its standard deviation $\sigma_i$, that is $\xi_i = x_i/\sigma_i $, $i=1,\ldots,m$.

\section{Training the sparse $\ell_1$-center classifier}
We next present an efficient and exact method for training a sparse $\ell_1$-center classifier.
We
 start by stating a  preliminary instrumental result, whose proof is reported in
the appendix Section~\ref{app:proofmedian}, and an ensuing definition.

\begin{proposition}[Weighted $\ell_1$ center]
\label{prop:wmedian}
Given 
a  real vector 
$z=(z_1, z_2,\ldots, z_p) $
and a nonnegative vector
$w=(w_1, \ldots, w_p) $, 
%
consider the weighted $\ell_1$ centering problem
\beq
d_w(z) \doteq \min_{\vartheta \in\Real{}} \sum_{i=1}^p w_i |z_i - \vartheta |.
\label{eq:weightedabsdev}
\eeq
Let 
\[
W(\zeta) \doteq  \sum_{\{i:\, z_i \leq \zeta\}} w_i,  \quad
\bar W \doteq \sum_{i = 1}^p w_i,
\]
and
\beq
\bar \zeta \doteq   \inf\{ \zeta : \;W(\zeta) \geq { \bar W}/{2} \}.
\label{eq:zeta_median}
\eeq
Then, an optimal solution  for problem \eqref{eq:weightedabsdev} is given by
\beq
\vartheta^* = \med_w (z) \doteq \dss \left\{ \ba{cl} \dss
\bar \zeta  & \mbox{if } W(\bar \zeta) >  \frac{ \bar W}{2}  \\ \\
\frac{1}{2}(\bar \zeta + \bar \zeta_+) &    \mbox{if } W(\bar \zeta) =  \frac{ \bar W}{2} ,
\ea\right.
\label{eq:thetastar_median}
\eeq
where $\bar \zeta_+ \doteq \min\{z_i, \, i=1,\ldots,p \, : z_i >  \bar \zeta\}$ is the smallest  element in $z$ that is strictly larger than
$\bar \zeta$. 
\est
\end{proposition}

\begin{definition}[Weighted median and dispersion]\rm 
Given 
a  row vector
$z$
and a nonnegative vector
$w$ of the same size, we define as the {\em weighted median} of $z$ the optimal solution of problem
\eqref{eq:weightedabsdev} given in \eqref{eq:thetastar_median}, and we denote it by $\med_w (z) $. We define as the {\em weighted median dispersion} the optimal value $d_w(z)$ of problem  \eqref{eq:weightedabsdev}.
We extend this notation to matrices, so that for a matrix $X\in\Real{m,n}$ we denote by
$\med_w (X) \in\Real{m} $ a vector whose $i$th component is $\med_w (X_{i,:}) $, 
where $X_{i,:}$ is the $i$th row of $X$, and we denote by $d_w(X)\in\Real{m}$ the vector of corresponding 
dispersions.
\est
\end{definition}

We now let $\calE$ and $\calD$ be defined as in Section~\ref{sec:L2train}, and we use the same notation as before for
$\theta ^\pm_\calD$, $ \theta^\pm_{\calE} $, $x_\calD$,  $x_{ \calE}$. 
%
Let then $J$ denote the objective to be minimized in \eqref{eq:L1obj}. 
For fixed index set $\calD$, we have that $J = J_\calD$, where
\beas
J_\calD  &=&  \frac{1}{n_+}\sum_{j\in\calJ^+}  \| x^{(j)} -  \theta ^+_\calD -  \theta_\calE\|_1
+
 \frac{1}{n_-}\sum_{j\in\calJ^-}  \|x^{(j)}-  \theta ^-_\calD -  \theta_\calE\|_1  \\
&=&  \frac{1}{n_+} \sum_{j\in\calJ^+}  \| (x^{(j)}_\calD -  \theta ^+_\calD) + (x^{(j)}_\calE -  \theta_\calE)\|_1
+
 \frac{1}{n_-}\sum_{j\in\calJ^-}  \|( x^{(j)}_\calD -  \theta ^-_\calD)  + (x^{(j)}_\calE  -  \theta_\calE)\|_1\\
&=& 
  \frac{1}{n_+}\sum_{j\in\calJ^+}  \| x^{(j)}_\calD -  \theta ^+_\calD\|_1
+   \frac{1}{n_+}\sum_{j\in\calJ^+}  \| x^{(j)}_\calE -  \theta_\calE\|_1 + 
 \frac{1}{n_-}\sum_{j\in\calJ^-}  \| x^{(j)}_\calD -  \theta ^-_\calD\|_1
 +
 \frac{1}{n_-} \sum_{j\in\calJ^-} \| x^{(j)}_\calE  -  \theta_\calE \|_1 \\
 &=&
  \frac{1}{n_+}\sum_{j\in\calJ^+}  \| x^{(j)}_\calD -  \theta ^+_\calD\|_1
+ 
 \frac{1}{n_-}\sum_{j\in\calJ^-}  \| x^{(j)}_\calD -  \theta ^-_\calD\|_1
 +
 \sum_{j=1}^n  w_j\| x^{(j)}_\calE  -  \theta_\calE \|_1,
\eeas
where
\[
 w_j = \left\{\ba{cl}\frac{1}{n_+} & \mbox{if } j\in \calJ^+ \\
 \frac{1}{n_-} & \mbox{if } j\in \calJ^-  \ea\right., \quad j=1,\ldots,n.
\]
We will next find the minimum of $J_\calD$ with respect to $\theta ^+_\calD$, $\theta ^-_\calD$ and
$\theta_\calE$. 
To this end, we observe that $J_\calD$ decouples as $J_\calD = \sum_{i=1}^m J_{\calD,i}$, where for $i=1,\ldots,m$,
\beq
J_{\calD,i} \doteq 
\left\{\ba{ll} \dss
 \frac{1}{n_+}\sum_{j\in\calJ^+}  | x^{(j)}_{i} -  \theta ^+_{i}|
+ 
 \frac{1}{n_-}\sum_{j\in\calJ^-}  | x^{(j)}_{i} -  \theta ^-_{i}|,
 & \mbox{if } i\in\calD \\ \dss
  \sum_{j=1}^n  w_j| x^{(j)}_{i}  -  \theta_{i} |,
  & \mbox{if } i\not \in\calD.
\ea\right. \label{eq:JDiL1}
%
\eeq
The minimum of $J_\calD$  is hence obtained by minimizing separately each component
$J_{\calD,i}$.
For $i\in\calD$, we have that 
 the optimal 
$\theta ^+_{i}$,  $\theta ^-_{i}$ are given by the (plain) medians  of the $x^{(j)}_{i} $ values in the positive and in the negative class, respectively, that is, recalling \eqref{eq:medianv},
\[
i\in\calD \quad \Rightarrow \quad 
\ba{rcl}
\theta ^{+*}_{i} &= & \mu^+_i \doteq  \med(\{x^{(j)}_i\}_{j\in\calJ^+})\\
\theta ^{-*}_{i} &= & \mu^-_i \doteq  \med(\{x^{(j)}_i\}_{j\in\calJ^-})
\ea
 \quad \Rightarrow \quad 
 J_{\calD,i}^* = d_{i}^+ + d_{i}^-,
\]
where $d^+$, $d^-$ are the vectors of median dispersions in the positive and negative class, respectively, whose components are, for $i=1,\ldots,m$,
\beq
\ba{lcr}\dss
d_i^+ &\doteq & \frac{1}{n_+}\sum_{j\in\calJ^+}  | x^{(j)}_{i} -  \mu^+_i| \\
\dss
d_i^- &\doteq & \frac{1}{n_-}\sum_{j\in\calJ^-}  | x^{(j)}_{i} -  \mu^-_i| .
\ea
\label{eq:meddisp}
\eeq
For $i\not\in\calD$, instead,  by 
observing that the entries of $ w$
in   \eqref{eq:JDiL1} are nonnegative, and applying Proposition~\ref{prop:wmedian}, we obtain that
the optimal solution
is the weighted median of {\em all} the observations, that is
\[
i\not\in\calD \quad \Rightarrow \quad 
\ba{rcl}
\theta ^{*}_{i} &= & \mu_i \doteq  \med_w(\{x^{(j)}_i\}_{j=1,\ldots,n})
\ea
 \quad \Rightarrow \quad 
 J_{\calD,i}^* = d_{i} ,
\]
where $d$ is the vector of weighted median dispersions over all the observations, whose components are, for $i=1,\ldots,m$,
\beq
d_i \doteq  \sum_{j=1}^n  w_j| x^{(j)}_{i}  -  \mu_{i} | = 
 \frac{1}{n_+}\sum_{j\in\calJ^+}| x^{(j)}_{i}  -  \mu_{i} |  +    \frac{1}{n_-}\sum_{j\in\calJ^-}| x^{(j)}_{i}  -  \mu_{i} | .
 \label{eq:wmeddisp}
\eeq
We are now in position to discuss how to optimize over the choice of the set $\calD$, that is how to decide which are the $k$ indices 
that should belong to $\calD$. First observe that
 $(d_{i}^+ + d_{i}^-) \leq d_i$, for all $i=1,\ldots,m$,
 since $d_i$ is the optimal value  of a minimization
that constrains $\theta_i^+$ to be equal to $\theta_i^-$, whereas $d_{i}^+ + d_{i}^-$ is the optimal value  of  the same minimization
without such constraint, and therefore its optimal objective value is no larger than $d_i$. 
Consider then the vector of differences
\[
e \doteq (d^+ + d^-) - d \leq 0. 
\]
The smallest (i.e., most negative) entry  in $e$ corresponds to an index $i$ for which it is maximally convenient (in terms of objective $J$ decrease)
choosing $i\in\calD$ rather than $i\not\in \calD$; the  second smallest entry  in $e$ corresponds to the second best choice, and so on.
The best $k$ indices to be included in $\calD$ are therefore those corresponding to the $k$ smallest entries
of vector $e$.
We have therefore proved the following
 
 \begin{proposition}
 \label{prop:ell1opt}
 The optimal solution of problem \eqref{eq:L1obj} is obtained as follows:
 \ben
 \item Compute the plain class medians
\beas
 \mu^+ & \doteq &  \med(\{x^{(j)}\}_{j\in\calJ^+})\\
 \mu^- & \doteq &  \med(\{x^{(j)}\}_{j\in\calJ^-})
 \eeas
 and the weighted median of all observations
 \[
  \mu \doteq  \med_w(\{x^{(j)}_i\}_{j=1,\ldots,n}),
 \]
 where the weight vector $w$ is such that,   for $j=1,\ldots,n$, $w_j = 1/n_+$ if $j\in\calJ^+$,
 and
 $w_j = 1/n_-$ if $j\in\calJ^-$.
 
 \item Compute the median dispersion vectors $d^+$, $d^-$ according to \eqref{eq:meddisp}, and the weighted median dispersion vector $d$ according to \eqref{eq:wmeddisp}, and compute the difference vector \[e \doteq   (d^+ + d^-) - d.\]

 \item Let $\calD$ be the set of the indices of the $k$ smallest elements in vector $e$, and let $\calE$ be the complementary index set.
 
 \item The optimal parameters $\theta ^+$, $\theta^-$ are given by
\beas
\theta ^+ &=& \mu^+_\calD +  \mu_\calE \\
\theta ^- &=& \mu^-_\calD +  \mu_\calE .
\eeas
 \een
 \end{proposition}

\begin{remark}[Numerical complexity for training the sparse $\ell_1$ classifier]\rm 
Computation of the medians in Step~1 of  Proposition~\ref{prop:ell1opt} 
can be performed with  in $O(m)$ operations, see, e.g., \cite{blum1973time}. Computation of the median dispersions requires $O(mn)$ operations, and finding the $k$ smallest elements in vector $e$ can be performed in $O(m\log k)$ operations, hence the whole procedure in Proposition~\ref{prop:ell1opt} is performed in $O(mn) + O(m\log k)$ operations.
Similar to the case discussed in Remark~\ref{rem:spacc}, also in the sparse $\ell_1$ center classifier one need to do a full ordering of an $m$-vector  only once  in order to obtain all the sparse classifiers for any sparsity level $k$.
\est
\end{remark}

\section{Experiments}
In this section, we perform an experimental evaluation of the proposed methods, comparing their performance with other feature selection techniques. The sparse $\ell_2$-center classifier is tested in the context of sentiment classification on text datasets. This is one of the most common application fields of the nearest centroid classifier. Instead, the sparse $\ell_1$-center classifier is evaluated on gene expression datasets. Since this type of data is usually affected by the presence of many outliers, the classifier with the $\ell_1$ distance criteria can be preferred over the $\ell_2$ version \cite{hall2009median}.
\subsection{Sparse $\ell_2$-center classifier}
We compared the proposed sparse $\ell_2$-center classifier with other feature selection methods for sentiment classification on text datasets. We considered three different datasets: the TwitterSentiment140 (TWTR) dataset, the MPQA Opinion Corpus Dataset, and the Stanford Sentiment Treebank (SST). Table \ref{tab:datasets} gives some details on the dataset sizes. Before classification, the dataset are preprocessed rescaling each feature by the inverse of its variance. Each dataset was randomly split in a training (80$\%$ of the dataset) and test (20$\%$ of the dataset) set. The results reported in this section are an average of 50 different random splits of the dataset.

\begin{table}[]
\caption{Text dataset sizes}
\label{tab:datasets}
\centering
\begin{tabular}{cccc}
\multicolumn{1}{l}{} & TWTR    & MPQA  & SST   \\ \hline
Number of features   & 273779  & 6208  & 16599 \\ \hline
Number of samples    & 1600000 & 10606 & 79654 \\ \hline
\end{tabular}
\end{table}

\begin{figure}
    \centering
    \includegraphics[width=0.4\textwidth]{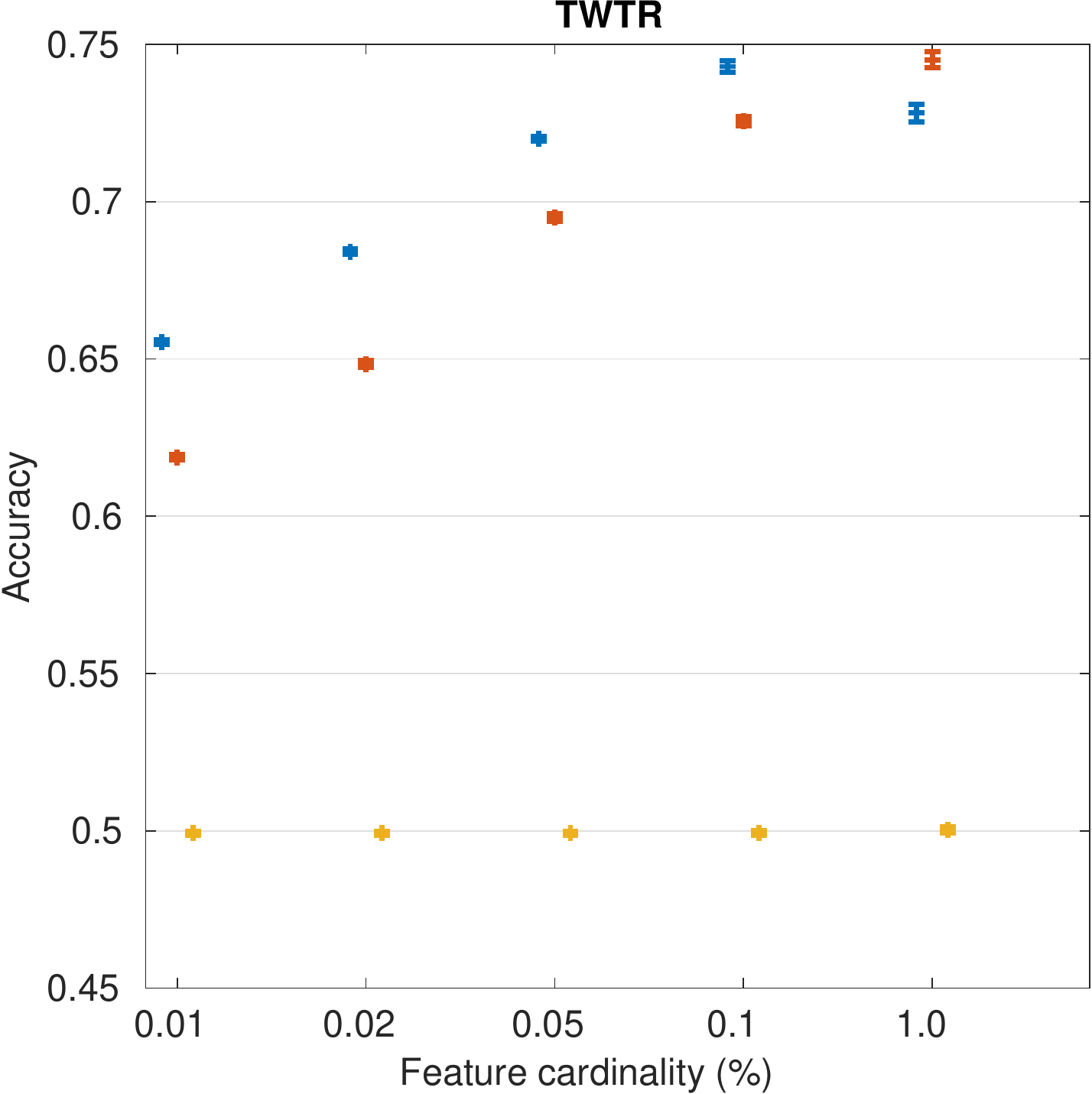}
    \includegraphics[width=0.522\textwidth]{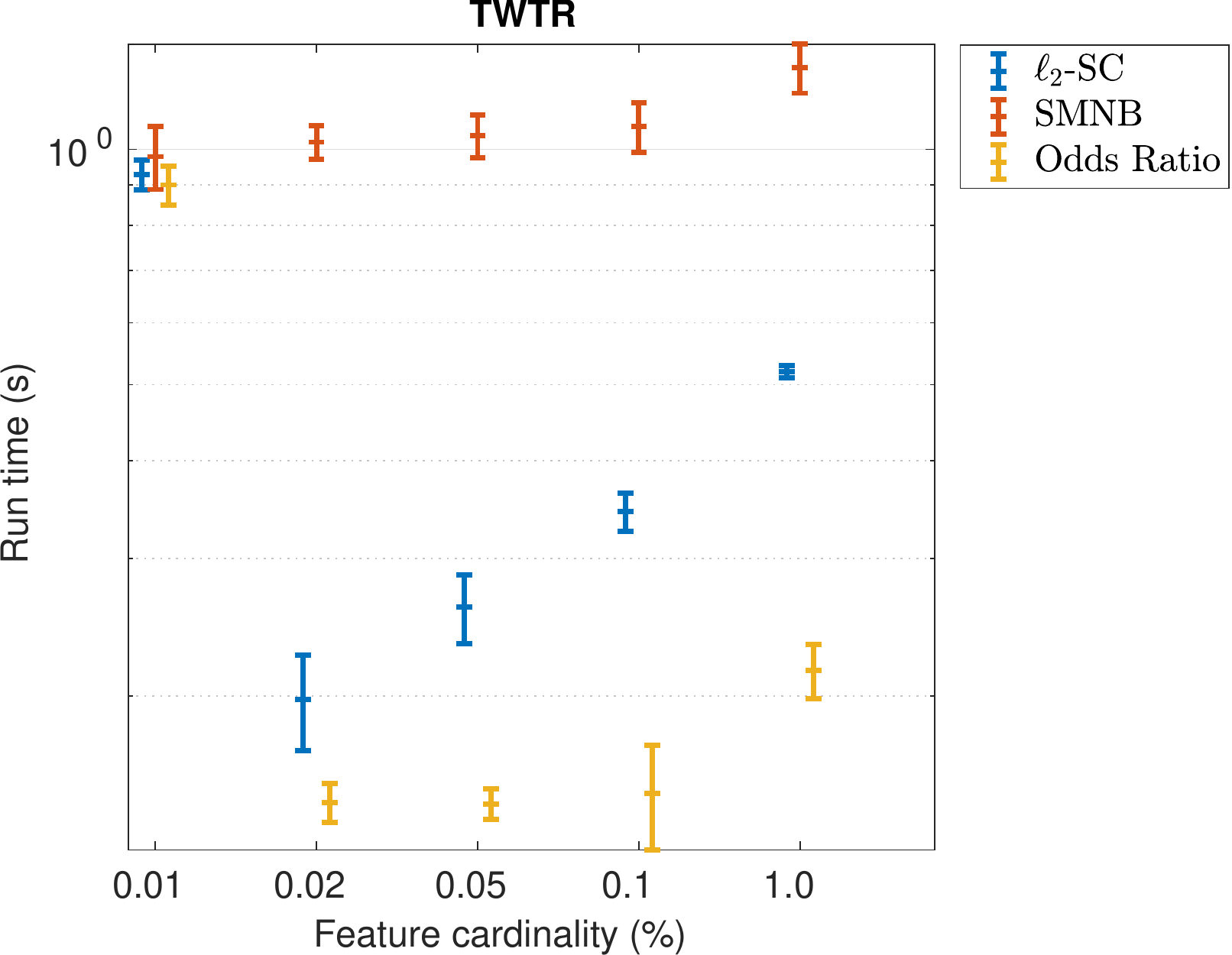}
    \includegraphics[width=0.4\textwidth]{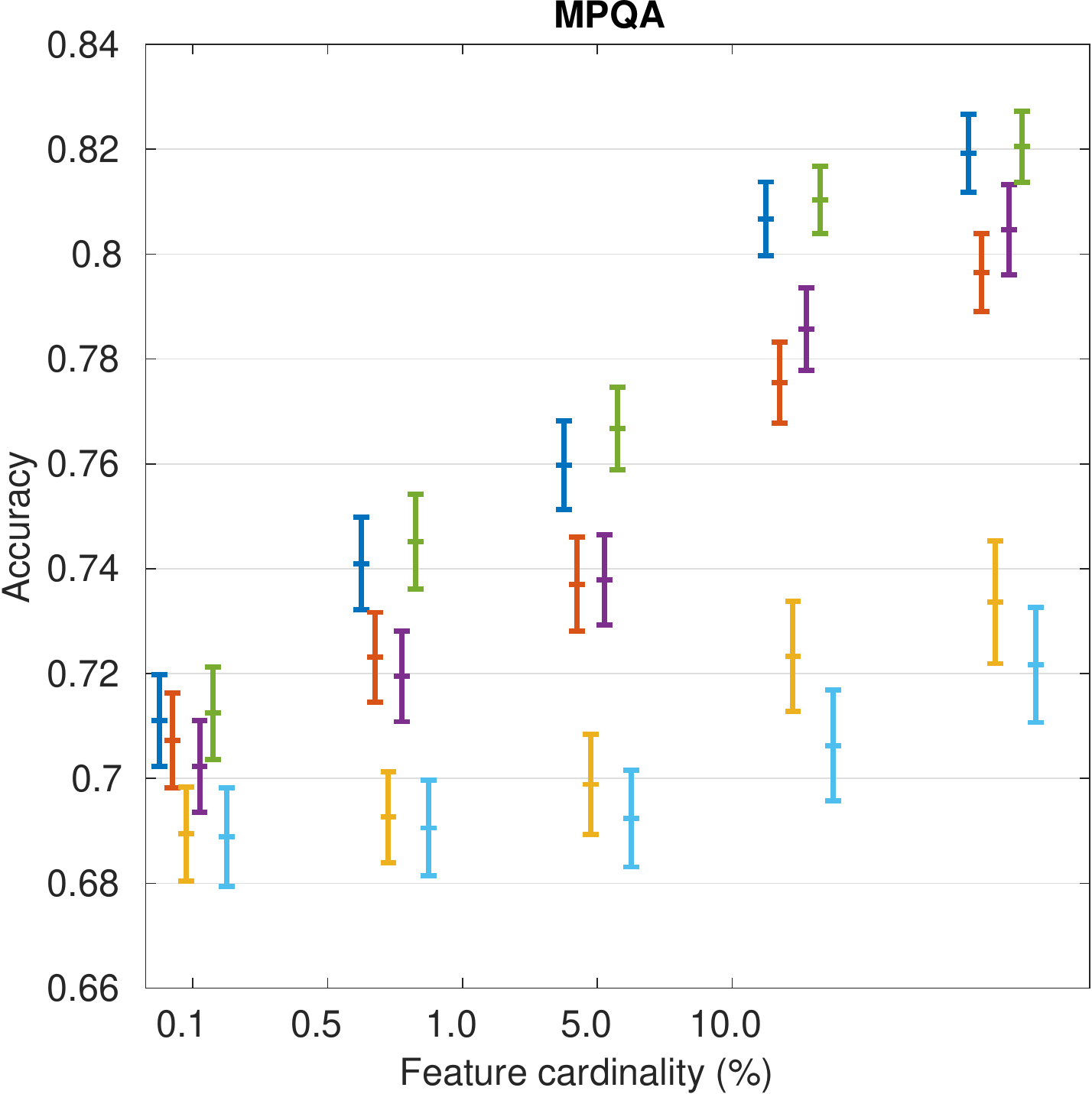}
    \includegraphics[width=0.54\textwidth]{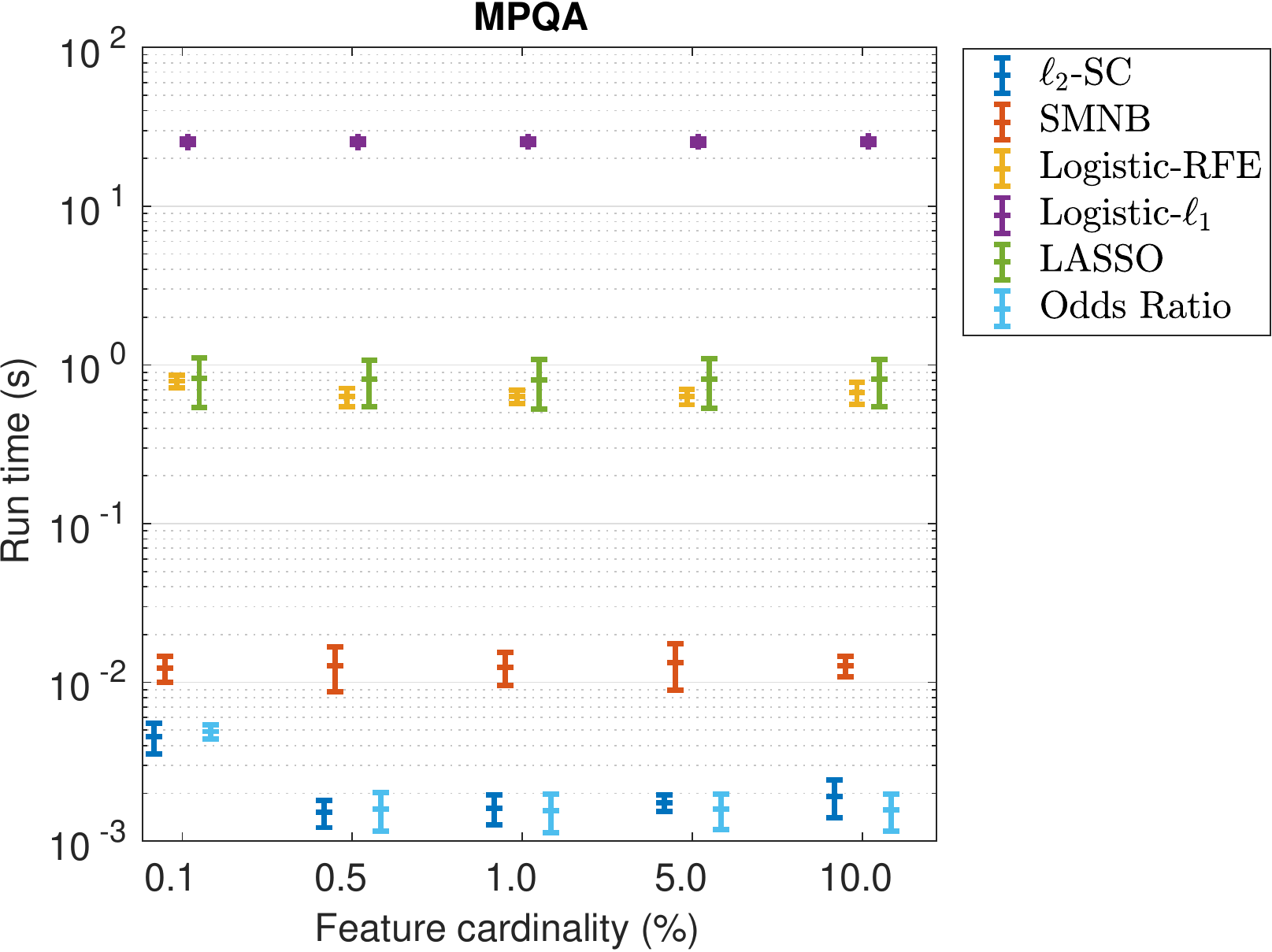}
    \includegraphics[width=0.4\textwidth]{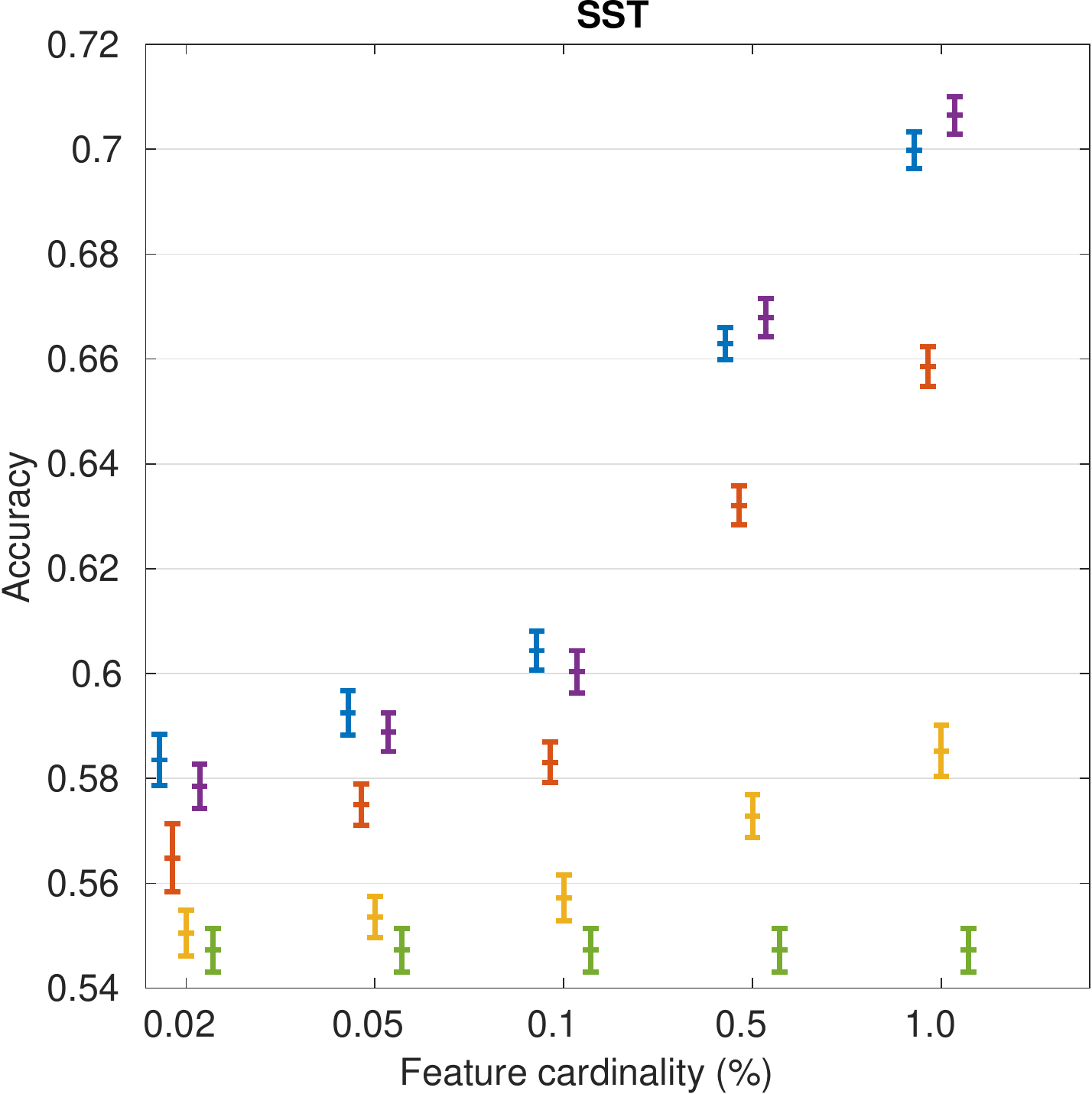}
    \includegraphics[width=0.54\textwidth]{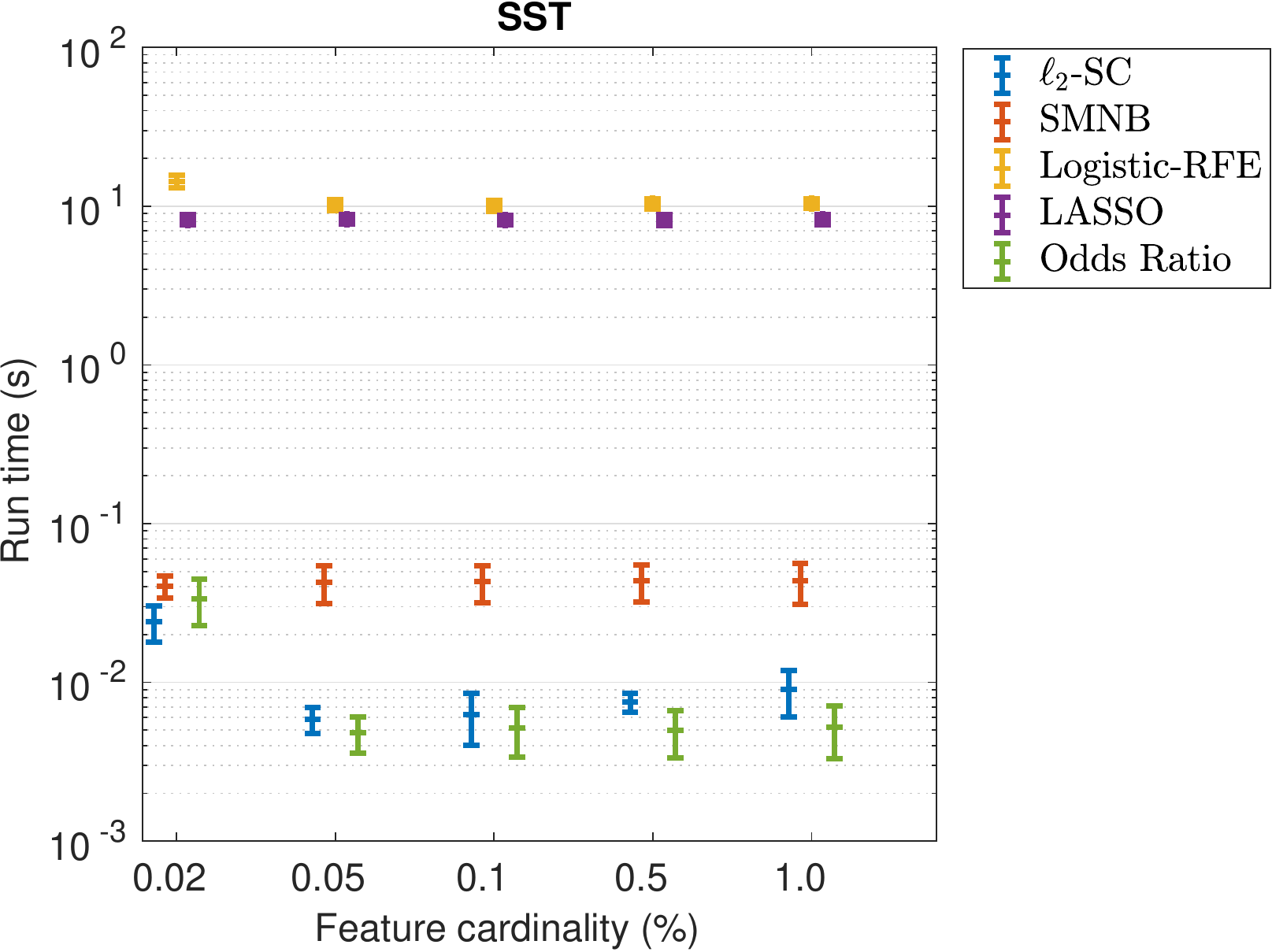}
    \caption{Classification accuracy and average run time.}
    \label{fig:acc_l2sc}
\end{figure}

For each dataset, we performed a two-stage classification procedure. In the first stage, we applied a feature selection method in order to reduce the number of features. Then, in the second stage we trained a classifier model, by employing only the selected features. In order to have a fair comparison, we used the same classifier for all the feature selection methods, namely a linear support vector machine classifier. 
We compared different feature selection methods: sparse $\ell_2$-centers ($\ell_2$-SC), sparse multinomial naive Bayes (SMNB), logistic regression with recursive feature selection (Logistic-RFE), $\ell_1$-regularized logistic regression (Logistic-$\ell_1$), Lasso, and Odds Ratio. Logistic-RFE, Logistic-$\ell_1$ and Lasso are not considered on some datasets, due to their high computational cost that makes them not viable when the dataset size is very large. 
Fig. \ref{fig:acc_l2sc} shows the accuracy performance and the average run time of the  different feature selection methods. These plots show that the sparse $\ell_2$-centers is competitive with other feature selection methods in terms of accuracy performance, while its run time is significantly lower than most of the other feature selection methods. The only method that has a comparable computational time is Odds Ratio, but its performance is  poor in terms of accuracy.

\subsection{Sparse $\ell_1$-center classifiers}
We compared the proposed sparse $\ell_1$-center classifier with other feature selection methods for RNA gene expression classification. We considered three datasets: Chin dataset \cite{chin2006genomic}, Chowdary dataset \cite{chowdary2006prognostic}, and Singh dataset \cite{singh2002gene}. The details of the datasets are summarized in Table \ref{tab:datasets_gene}. As done in the $\ell_2$ case, we subdivided each dataset in a training (80$\%$ of the dataset) and test (20$\%$ of the dataset) set, and we tested 50 random splits.
\begin{table}[]
\caption{RNA gene expression dataset sizes}
\label{tab:datasets_gene}
\centering
\setlength\tabcolsep{3pt} 
\begin{tabular}{cccc}
\multicolumn{1}{l}{} & Chowdary  & Chin  & Singh \\
\multicolumn{1}{l}{} & (Breast Cancer)   & (Breast Cancer)  & (Prostate Cancer)    \\ \hline
N. features   & 22283  & 22215 & 12600 \\ \hline
N. samples    & 104 & 118 & 102 \\ \hline
\end{tabular}
\end{table}

\begin{figure}
    \centering
    \includegraphics[width=0.4\textwidth]{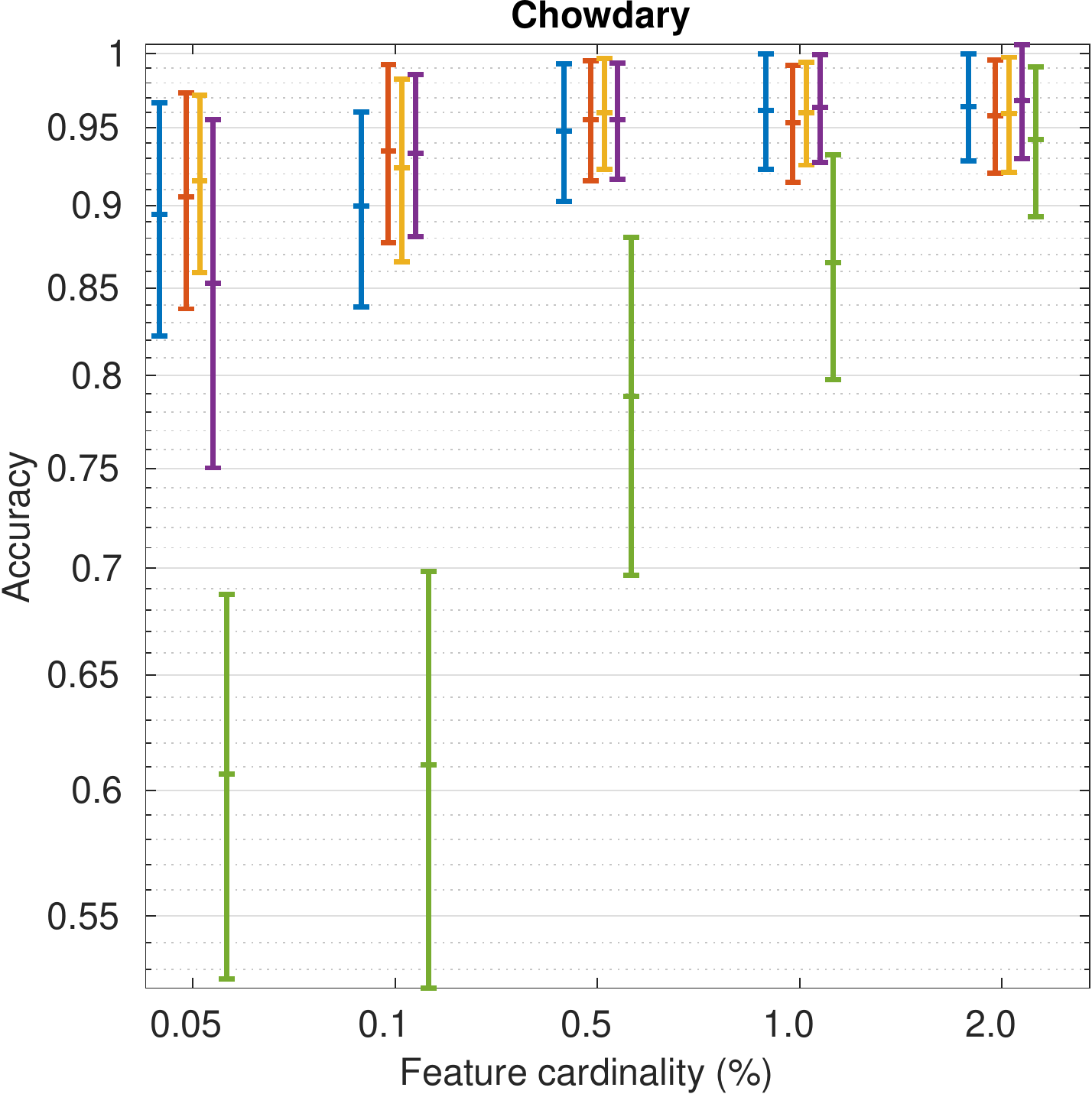}
    \includegraphics[width=0.522\textwidth]{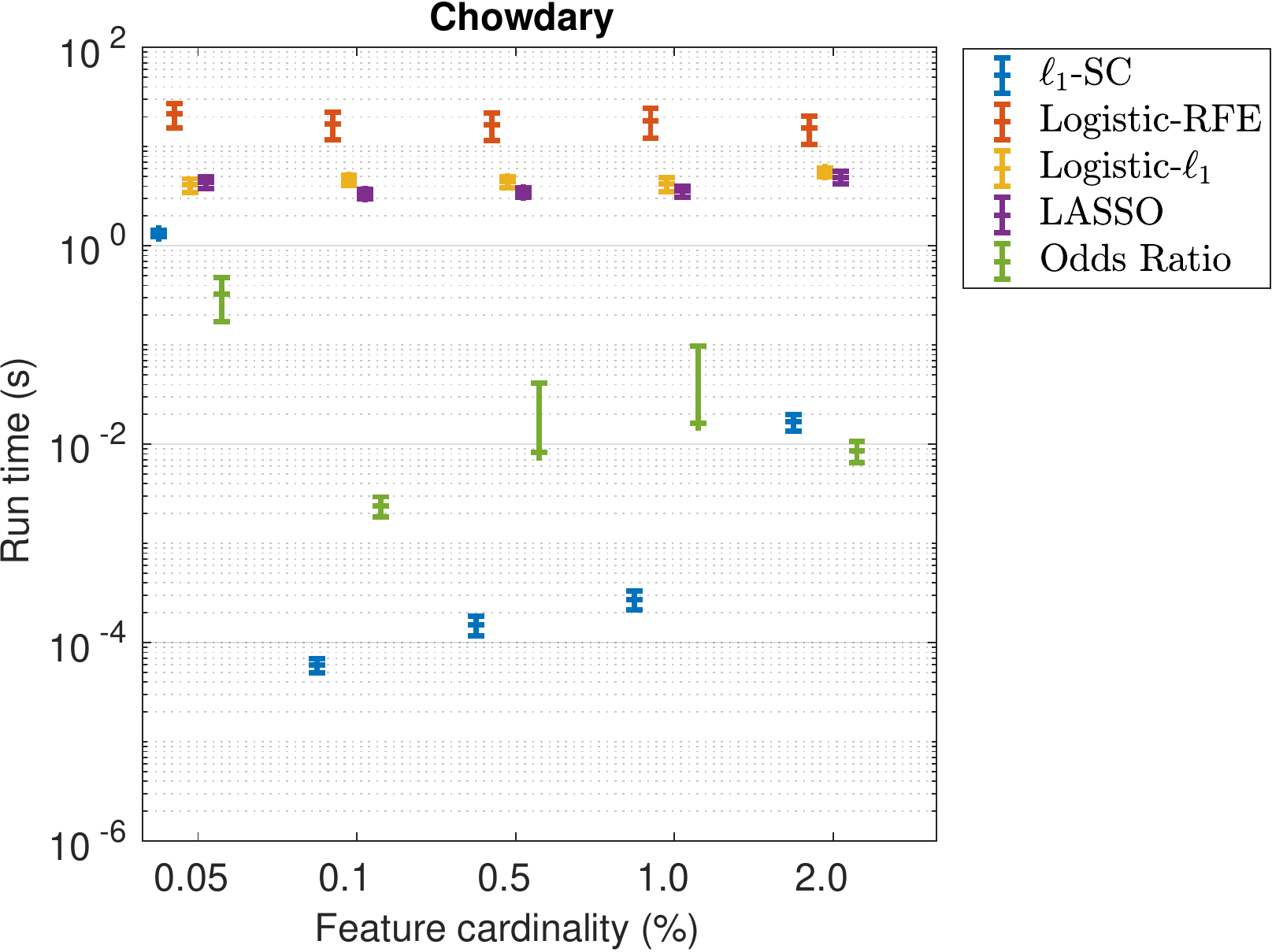}
    \includegraphics[width=0.4\textwidth]{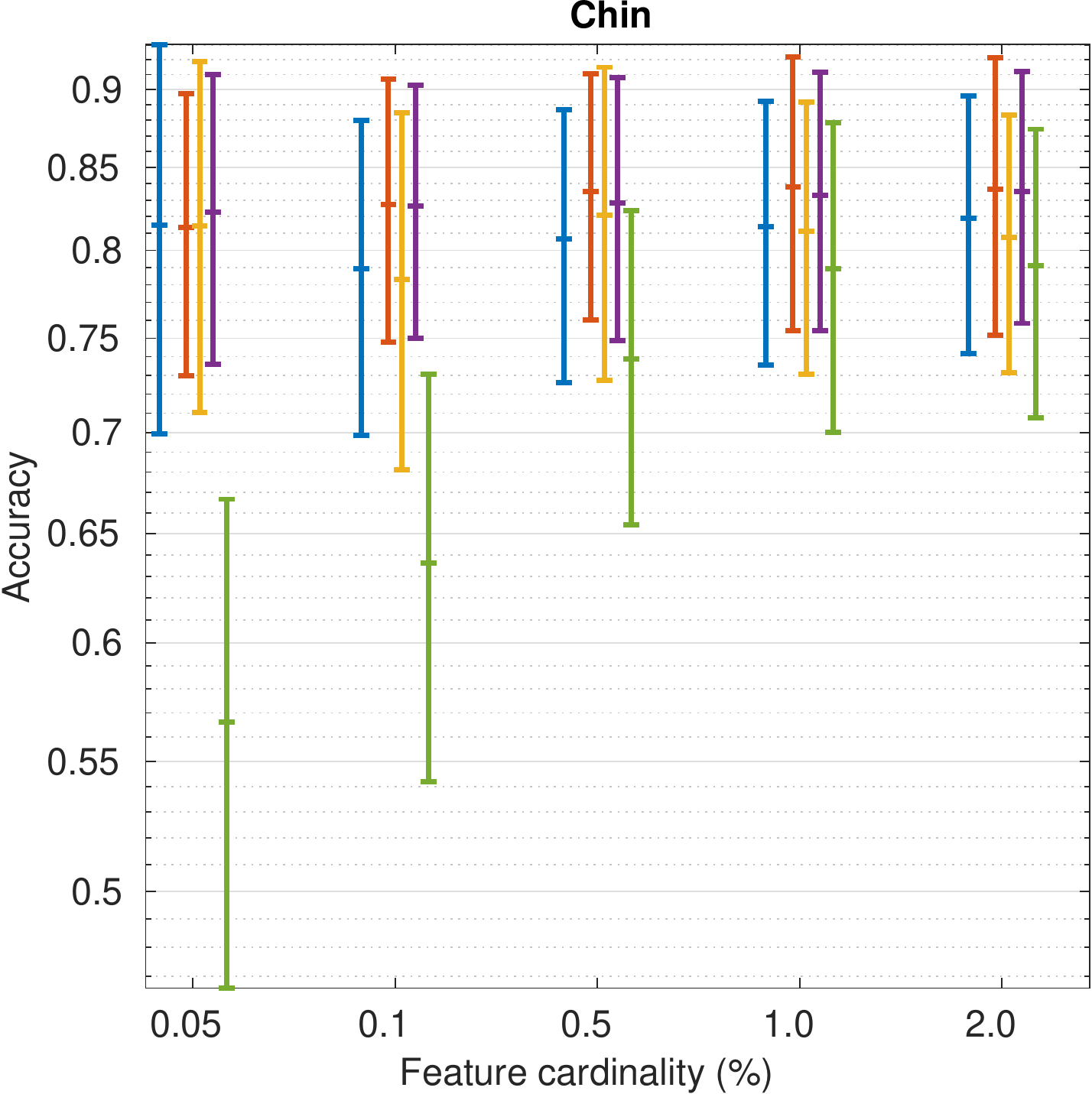}
    \includegraphics[width=0.54\textwidth]{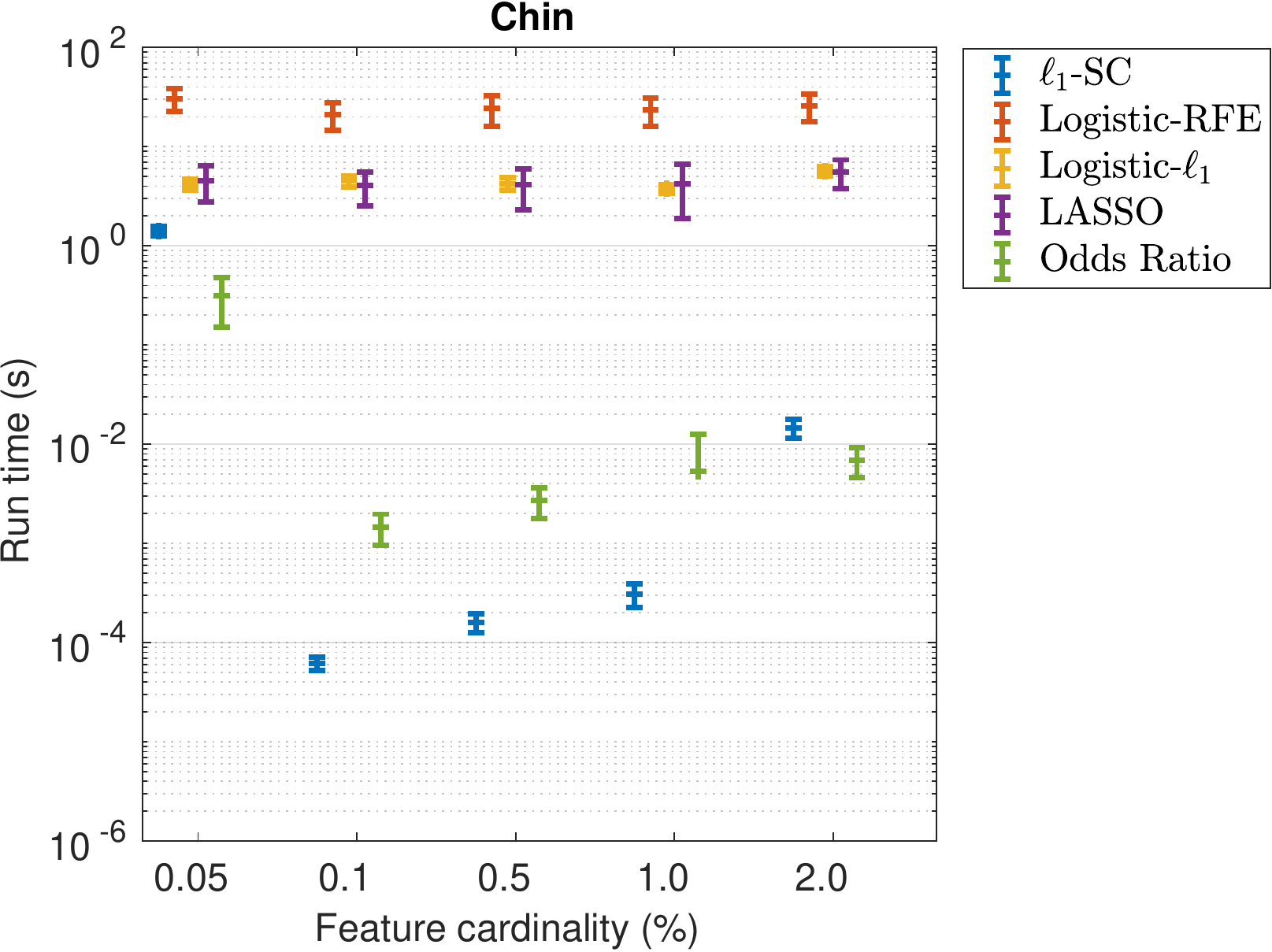}
    \includegraphics[width=0.4\textwidth]{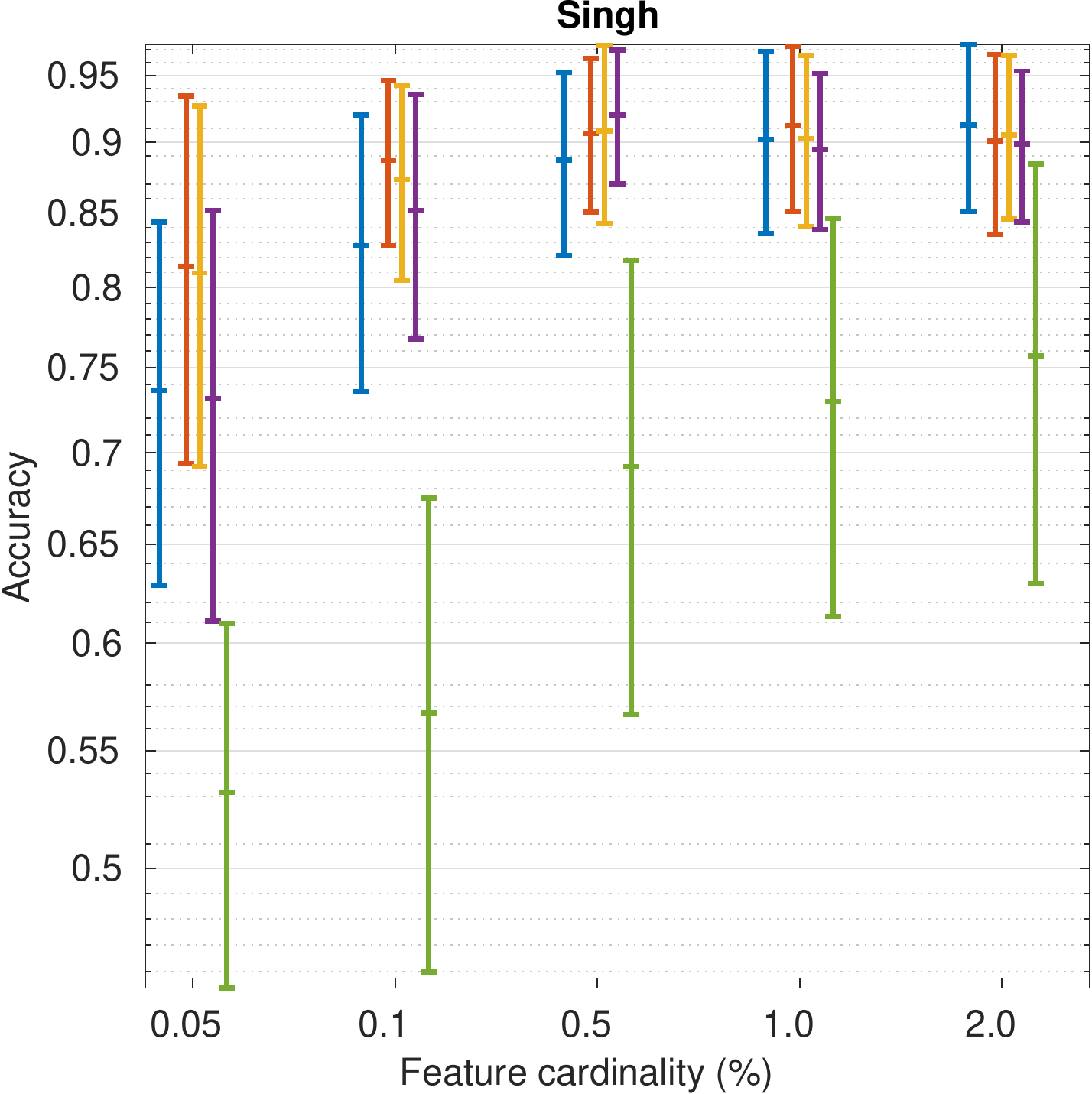}
    \includegraphics[width=0.54\textwidth]{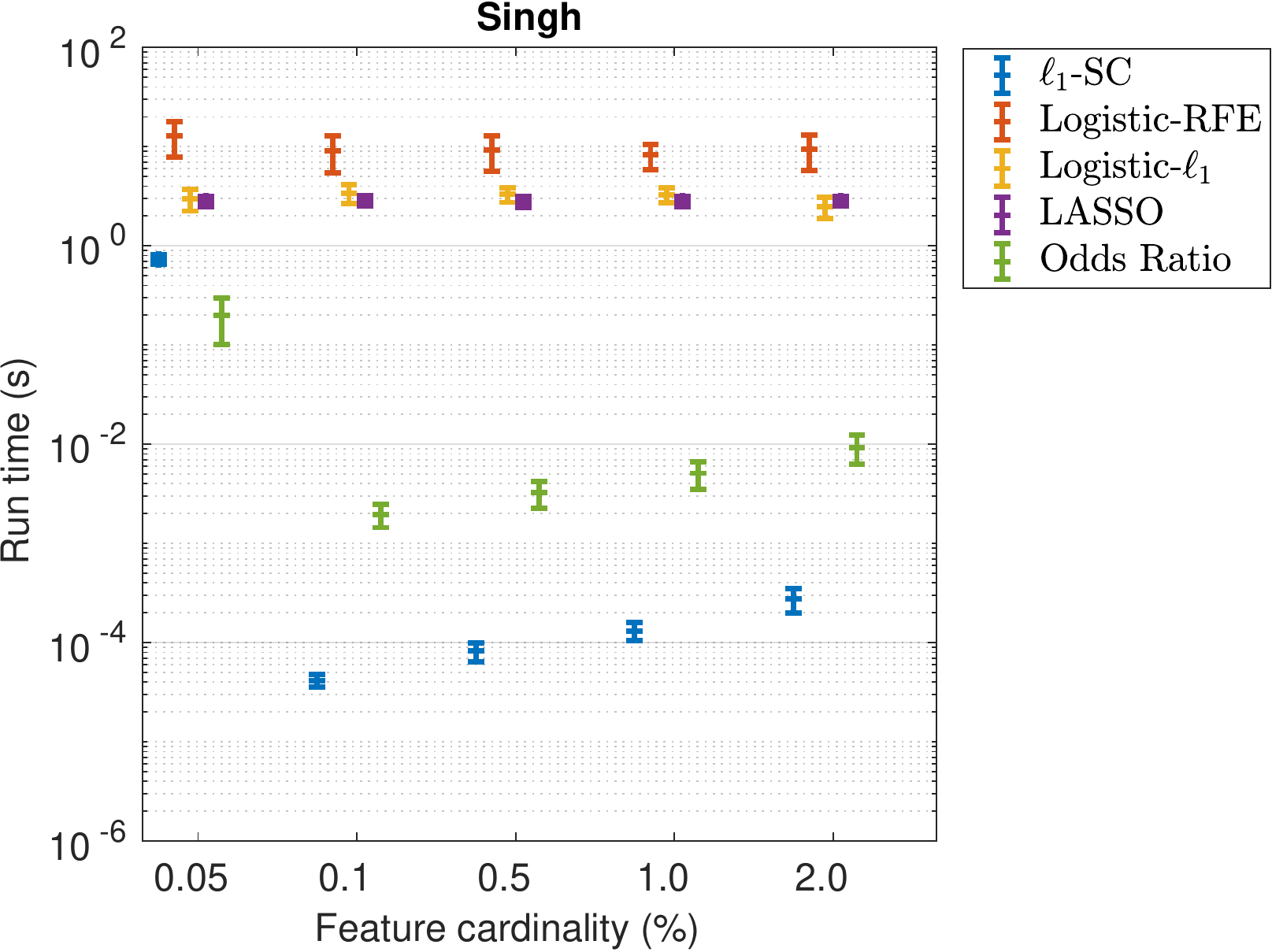}
    \caption{Classification accuracy and average run time.}
    \label{fig:acc_l1sc}
\end{figure}

For each dataset, we performed a two-stage procedure, as explained in the previous section. In the first stage, we compared five feature selection methods: sparse $\ell_1$-centers ($\ell_1$-SC), $\ell_1$-regularized logistic regression (Logistic-$\ell_1$), logistic regression with recursive feature elimination (Logistic-RFE), Lasso, and Odds Ratio. Sparse Multinomial Naive Bayes (SMNB) is not taken into account in this experiment since the gene expression datasets can have negative features and SMNB can only be applied to datasets with positive features.  In the second stage, we used a linear SVM classifier, as in the previous section. Figure \ref{fig:acc_l1sc} shows the balanced accuracy and average run time of the feature selection methods. Also in this experiment we  observe that the proposed method provides an accuracy performance which is similar to that of state-of-the-art techniques, but with a significantly lower computational time.

\section{Conclusion}
In this paper we proposed two types of  sparse center classifiers, based respectively on $\ell_1$ and the $\ell_2$ distance metrics. The proposed methods perform simultaneous classification and feature selection, and  in both cases the proposed training method selects the optimal set of features in a quasi-linear computing time.
 The experimental results also show that the proposed methods achieve accuracy   levels that are on par with state-of-the-art feature selection methods, while being substantially faster.

\section{Appendix}

\subsection{Proof of Proposition~\ref{prop:wmedian}}
\label{app:proofmedian}
Let $\tilde w \doteq w /\bar W$. Since
$\tilde w \geq 0$ and $\sum_{i=1}^p \tilde w_i = 1$, it can be interpreted as the probability  distribution of a
discrete random variable $Z$  with support in $z_{ 1},\ldots, z_{ p}$, and corresponding probability mass
$\tilde w_{ 1},\ldots, \tilde w_{ p}$. Note that values in vector $z$ may be repeated, in which case the probability mass relative to a repeated support point  is the sum of the corresponding probability values in vector $\tilde w$. With such stochastic interpretation, the objective in 
\eqref{eq:weightedabsdev} can be written in terms of the expectation 
$
\E\{|Z - \vartheta|\}
$, and then the problem becomes
\beq
d_w(z) = \bar W \min_{\vartheta \in\Real{}} \E\{|Z - \vartheta|\}.
\label{eq:weightedabsdev_prf}
\eeq
When $Z$ has an absolutely continuous distribution, it is well known (see, e.g., \cite{haldane1948note}) that the value $\vartheta^*$ that minimizes the absolute expected loss is the {\em median} of the probability distribution of $Z$, that is, the 0.5 quantile of the distribution.
In the case of a discrete probability distribution, the definition of  median is any value $\mu$ such that
\beq
\prob\{Z\leq \mu\} \geq \frac{1}{2}, \quad \mbox{and} \quad \prob\{Z\geq \mu\} \geq \frac{1}{2}.
\label{eq:def:median}
\eeq
Now, suppose that $\mu$ is a median for our discrete random variable $Z$, and consider any given $\vartheta > \mu$. 
If $Z \leq \mu$, then $|Z- \mu|  = \mu-Z$ and  since $ \mu < \vartheta$  we also have $Z < \vartheta$ whence
 $ |Z- \vartheta| = \vartheta - Z$, and therefore
 \[
  |Z- \vartheta| - |Z- \mu| =  (\vartheta - Z)    -  (\mu - Z) =  \vartheta - \mu, \quad \mbox{for } Z \leq \mu.
 \] 
 If instead $Z > \mu$, then 
\beas
  |Z- \vartheta| - |Z- \mu| &=& |Z- \vartheta| - (Z- \mu) =   |Z- \mu + \mu -\vartheta| - (Z- \mu) \\
  & \geq &  |Z- \mu| - |\mu -\vartheta| - (Z- \mu) = (Z-\mu) - (\vartheta-\mu) -   (Z- \mu)  \\
  &= & - ( \vartheta-\mu) , \quad \mbox{for } Z > \mu.
 \eeas
Therefore, for any given $\vartheta > \mu$, we have that
\beas
\E\{|Z- \vartheta| - |Z- \mu| \} & \geq &   (\vartheta - \mu) \prob\{Z\leq \mu\} -  (\vartheta - \mu) \prob\{Z> \mu\} 
\\ &=&  (\vartheta - \mu) \left( \prob\{Z\leq \mu\}     -  \prob\{Z> \mu\} \right)  \\
&=&  (\vartheta - \mu) \left( 2\prob\{Z\leq \mu\}     -  1\} \right)
\\ &\geq & 0, \quad \mbox{for all }\vartheta > \mu.
\eeas
where the last inequality follows from the fact that $\mu$ is a distribution median and hence from the definition
in \eqref{eq:def:median} it holds that  $\prob\{Z\leq \mu\} \geq 1/2$.
The whole reasoning can be repeated symmetrically for any given $\vartheta < \mu$, obtaining
\beas
  |Z- \vartheta| - |Z- \mu| &\geq &  - ( \mu- \vartheta) , \quad \mbox{for } Z < \mu, \\
  |Z- \vartheta| - |Z- \mu| &=  &   ( \mu- \vartheta) , \quad \mbox{for } Z \geq  \mu.
 \eeas
 Then again
 \beas
\E\{|Z- \vartheta| - |Z- \mu| \} & \geq &  - ( \mu- \vartheta) \prob\{Z< \mu\} +  ( \mu- \vartheta) \prob\{Z\geq  \mu\} 
\\ &=& ( \mu- \vartheta) \left( \prob\{Z\geq \mu\}     -  \prob\{Z< \mu\} \right)  \\
&=&   ( \mu- \vartheta) \left( 2\prob\{Z\geq \mu\}     -  1\} \right)
\\ &\geq & 0, \quad \mbox{for all }\vartheta < \mu,
\eeas
where the last inequality follows from the fact that $\mu$ is a distribution median and hence from the definition
in \eqref{eq:def:median} it holds that  $\prob\{Z\geq \mu\} \geq 1/2$. Putting things together, we have that
\beas
\E\{|Z- \vartheta|\} - \E\{|Z- \mu| \} &=& \E\{|Z- \vartheta| - |Z- \mu| \}  \geq 0,
\quad \forall \vartheta,
\eeas
which implies that the minimum of $\E\{|Z- \mu| \} $  is attained at $\vartheta = \mu$, where $\mu$ is a median of the distribution.

We next conclude the proof by showing that $\vartheta^*$ in \eqref{eq:thetastar_median} is indeed a median, in the sense of definition
\eqref{eq:def:median}. Observe first that 
$W(\zeta) \doteq  \sum_{i: z_i \leq \zeta} w_i$
is proportional to  the cumulative distribution function of $Z$, that is
\[
W(\zeta) = \bar W \tilde W(\zeta),\quad \tilde W(\zeta)\doteq \prob\{Z \leq \zeta\},
\] 
and that 
\eqref{eq:zeta_median} implies that $\tilde W(\bar \zeta) \geq 1/2$, and $\tilde W(\zeta ) < 1/2$ for all $\zeta < \bar \zeta$.
Also, since by definition of $ \bar \zeta_+$ no probability mass is present in the interior of the interval $[\bar \zeta, \bar \zeta_+]$, we have from \eqref{eq:thetastar_median} that $\tilde W (\vartheta^*) \equiv \tilde  W(\bar \zeta) $.
Then,   from \eqref{eq:thetastar_median} it follows immediately that $\prob\{Z \leq \vartheta^*\} = \tilde W(\vartheta^*)  \equiv  \tilde W(\bar \zeta)  \geq 1/2$,
which shows that $\vartheta^*$ satisfies the condition on the left in \eqref{eq:def:median}.
We next analyze  the condition on the right in \eqref{eq:def:median}, which concerns verifying that
$\prob\{Z \geq \vartheta^*\}\geq 1/2$. To this purpose, we distinguish two cases: 
case (a), where  $\tilde W(\vartheta^*) >1/2$,
and case (b),  where $\tilde W(\vartheta^*) =1/2$. 
In case (a), we have $\vartheta^* \equiv \bar \zeta$ and hence, as discussed above, $\tilde W(\zeta) < 1/2$ for all $\zeta <\vartheta^*$, which implies that $\prob\{Z < \vartheta^*\} < 1/2$ (while $\prob\{Z \leq \vartheta^*\} \geq 1/2$, since there is a positive probability mass
at $\vartheta^*$), and therefore
\[
\prob\{Z \geq \vartheta^*\} = 1 - \prob\{Z < \vartheta^*\} > 1/2.
\]
In case (b), we have instead
\beas
\prob\{Z \geq \vartheta^*\} &=& \prob\{Z = \vartheta^*\} + \prob\{Z >\vartheta^*\} \\ &=& 
 \prob\{Z = \vartheta^*\} + 1 - \prob\{Z \leq \vartheta^*\} =  \prob\{Z = \vartheta^*\}  + 1/2\\ & =& 1/2 ,
\eeas
where the last equality follows from the fact that in case (b) we have $\prob\{Z = \vartheta^*\} = 0$, since 
$ \vartheta^*$ is the mid point of the  interval $[\bar \zeta, \bar \zeta_+]$, in the interior of which there is no probability mass,
by construction.
 \qed


\begin{thebibliography}{10}

\bibitem{askari2019naive}
A.~Askari, A.~d'Aspremont, and L.~E. Ghaoui.
\newblock Naive feature selection: Sparsity in naive bayes.
\newblock {\em arXiv preprint arXiv:1905.09884}, 2019.

\bibitem{blum1973time}
M.~Blum, R.~W. Floyd, V.~R. Pratt, R.~L. Rivest, and R.~E. Tarjan.
\newblock Time bounds for selection.
\newblock {\em J. Comput. Syst. Sci.}, 7(4):448--461, 1973.

\bibitem{chin2006genomic}
K.~Chin, S.~DeVries, J.~Fridlyand, P.~T. Spellman, R.~Roydasgupta, W.-L. Kuo,
  A.~Lapuk, R.~M. Neve, Z.~Qian, T.~Ryder, et~al.
\newblock Genomic and transcriptional aberrations linked to breast cancer
  pathophysiologies.
\newblock {\em Cancer cell}, 10(6):529--541, 2006.

\bibitem{chowdary2006prognostic}
D.~Chowdary, J.~Lathrop, J.~Skelton, K.~Curtin, T.~Briggs, Y.~Zhang, J.~Yu,
  Y.~Wang, and A.~Mazumder.
\newblock Prognostic gene expression signatures can be measured in tissues
  collected in rnalater preservative.
\newblock {\em The journal of molecular diagnostics}, 8(1):31--39, 2006.

\bibitem{haldane1948note}
J.~Haldane.
\newblock Note on the median of a multivariate distribution.
\newblock {\em Biometrika}, 35(3-4):414--417, 1948.

\bibitem{hall2009median}
P.~Hall, D.~Titterington, and J.-H. Xue.
\newblock Median-based classifiers for high-dimensional data.
\newblock {\em Journal of the American Statistical Association},
  104(488):1597--1608, 2009.

\bibitem{han2000centroid}
E.-H.~S. Han and G.~Karypis.
\newblock Centroid-based document classification: Analysis and experimental
  results.
\newblock In {\em European conference on principles of data mining and
  knowledge discovery}, pages 424--431. Springer, 2000.

\bibitem{manning2008vector}
C.~Manning, P.~Raghavan, and H.~Sch{\"u}tze.
\newblock Vector space classification.
\newblock {\em Introduction to Information Retrieval}, 2008.

\bibitem{mladenic1999feature}
D.~Mladenic and M.~Grobelnik.
\newblock Feature selection for unbalanced class distribution and naive bayes.
\newblock In {\em ICML}, volume~99, pages 258--267, 1999.

\bibitem{ng2004feature}
A.~Y. Ng.
\newblock Feature selection, l 1 vs. l 2 regularization, and rotational
  invariance.
\newblock In {\em Proceedings of the twenty-first international conference on
  Machine learning}, page~78. ACM, 2004.

\bibitem{singh2002gene}
D.~Singh, P.~G. Febbo, K.~Ross, D.~G. Jackson, J.~Manola, C.~Ladd, P.~Tamayo,
  A.~A. Renshaw, A.~V. D'Amico, J.~P. Richie, et~al.
\newblock Gene expression correlates of clinical prostate cancer behavior.
\newblock {\em Cancer cell}, 1(2):203--209, 2002.

\bibitem{tibshirani1996regression}
R.~Tibshirani.
\newblock Regression shrinkage and selection via the lasso.
\newblock {\em Journal of the Royal Statistical Society: Series B
  (Methodological)}, 58(1):267--288, 1996.

\bibitem{tibshirani2002diagnosis}
R.~Tibshirani, T.~Hastie, B.~Narasimhan, and G.~Chu.
\newblock Diagnosis of multiple cancer types by shrunken centroids of gene
  expression.
\newblock {\em Proceedings of the National Academy of Sciences},
  99(10):6567--6572, 2002.

\end{thebibliography}
 \end{document}